\documentclass[letter, 10pt, conference]{ieeeconf}
\IEEEoverridecommandlockouts                              
\usepackage{cite}
\usepackage{algorithm}
\usepackage{algpseudocode}
\usepackage{amsmath}
\usepackage{amsfonts}
\usepackage{amssymb}  
\usepackage[utf8]{inputenc}
\usepackage[english]{babel}
\usepackage[T1]{fontenc}
\usepackage{url}
\usepackage{soul}
\usepackage{color}
\usepackage{graphicx}
\usepackage{xcolor}
\usepackage[export]{adjustbox}
\usepackage{subcaption}

\newcommand{\ours}{\textit{BeBOP}}

\begin{document}
\bstctlcite{IEEEexample:BSTcontrol}
\title{\LARGE 
BeBOP -- Combining Reactive Planning and Bayesian Optimization to Solve Robotic Manipulation Tasks}

\author{Jonathan Styrud\authorrefmark{1}\authorrefmark{3}, Matthias Mayr\authorrefmark{2}, Erik Hellsten\authorrefmark{2}, Volker Krueger\authorrefmark{2} and Christian Smith\authorrefmark{3}
\thanks{This project is supported by the Wallenberg AI, Autonomous Systems, and Software Program (WASP) funded by the Knut and Alice Wallenberg Foundation. The authors gratefully acknowledge this support. We also want to thank the authors of \cite{nasiriany2022augmenting} for promptly answering all of our questions.}
\thanks{\authorrefmark{1}ABB Robotics, Västerås, Sweden}
\thanks{\authorrefmark{2}Lund University, Lund, Sweden}
\thanks{\authorrefmark{3}Division of Robotics, Perception and Learning, Royal Institute of Technology (KTH), Stockholm, Sweden}}

\maketitle

\begin{abstract}
Robotic systems for manipulation tasks are increasingly expected to be easy to configure for new tasks. While in the past, robot programs were often written statically and tuned manually, the current, faster transition times call for robust, modular and interpretable solutions that also allow a robotic system to learn how to perform a task.
We propose the method Behavior-based Bayesian Optimization and Planning (\ours)~that combines two approaches for generating behavior trees: we build the structure using a reactive planner and learn specific parameters with Bayesian optimization. The method is evaluated on a set of robotic manipulation benchmarks and is shown to outperform state-of-the-art reinforcement learning algorithms by being up to 46 times faster while simultaneously being less dependent on reward shaping. We also propose a modification to the uncertainty estimate for the random forest surrogate models that drastically improves the results.
\end{abstract}

\begin{keywords}
Behavior Trees, Bayesian Optimization, Task Planning, Robotic manipulation
\end{keywords}

\section{Introduction}
Modern robots are capable of solving complex tasks in controlled environments with high reliability and precision. However, recent trends are pointing towards smaller product batches and more frequent updates of robot programs. At the same time, the market share of collaborative robots is growing steadily, while workspaces shared with humans makes for more unpredictable environments. As a result, it is becoming increasingly important to create new robot policies or programs quickly without the need for advanced programming skills and for those programs to be reactive to changes in the environment.
There are two main groups of methods to generate policies automatically, both with their own advantages and drawbacks. Firstly, automated planners~\cite{ghallab_automated_2016} can be very efficient, but require that the planning domain is modelled sufficiently well. As an example, a planner can only avoid obstacles that are represented in the domain. Planners also tend not to scale well to higher task complexity. The second group, colloquially known as Machine Learning (ML), typically builds a model by interacting with the environment and is thus not limited by preexisting knowledge. There are also cases where ML methods scale better than planners, as they can use a probabilistic approach instead of an exhaustive approach. However, the learning is often not very efficient and for smaller tasks, an automated planner can be many orders of magnitude faster. This hampers the use of ML-based methods as even state-of-the-art methods can take hours to days of interaction time to learn a new task. As an example, the \textit{MAPLE} runs for the benchmarks in these paper takes several days on a normal workstation to learn.
Another considerable drawback is that many of the ML algorithms, often in the Reinforcement Learning (RL) subgroup, are designed to use neural networks that are known to lack the transparency and modularity of other architectures.

An increasingly popular alternative in robotics is to instead represent the policy with Behavior Trees (BTs)~\cite{iovino_survey_2022, colledanchise_behavior_2018}. The main advantages are that BTs have explicit support for task hierarchy, action sequencing, reactivity and they are inherently modular. They are also transparent and readable, which enables manual and automated analysis and validation~\cite{colledanchise_how_2017} as well as manual editing. Those are strong advantages when compared to neural networks, especially in an industrial setting.

\begin{figure}[tpb!]
    \setlength{\fboxrule}{0pt}
		\framebox{\parbox{3in}{
            \centering
            \includegraphics[width=0.48\textwidth]{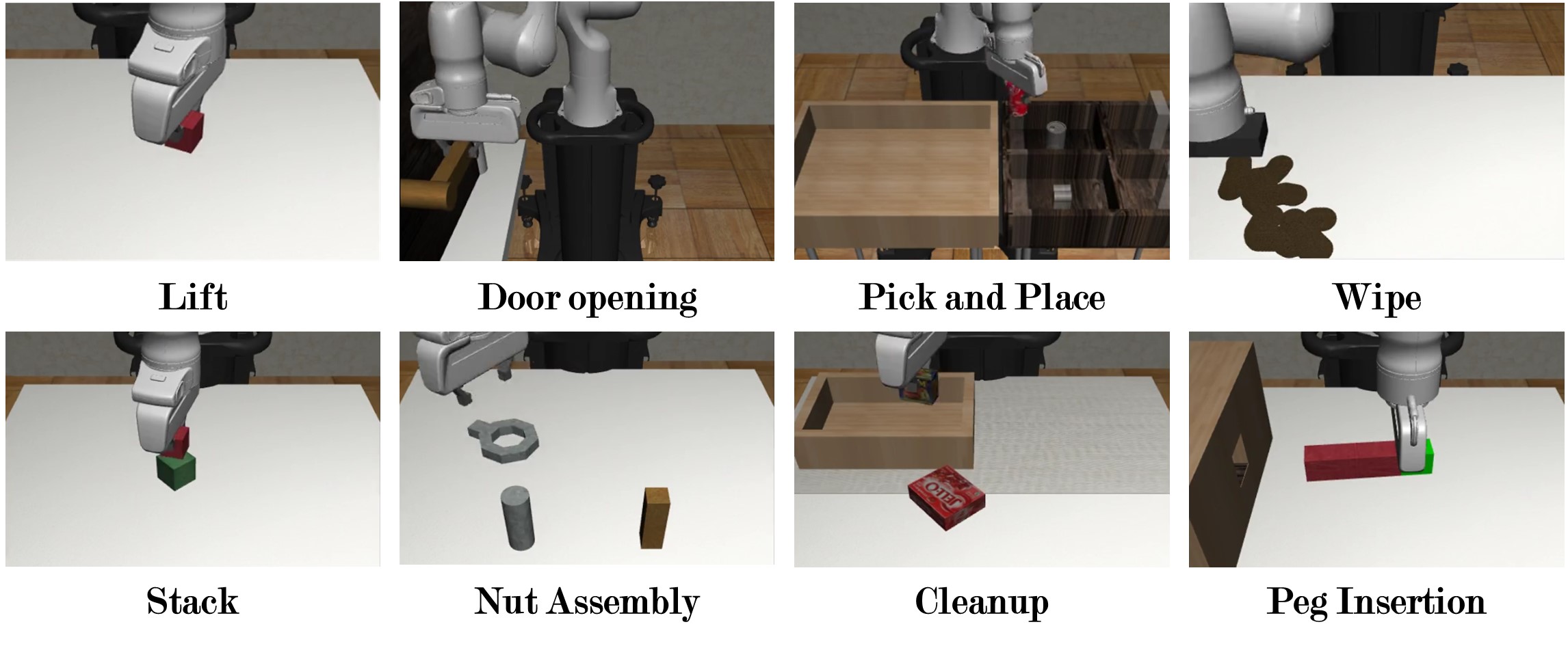}
        }
    }
\caption{The eight simulation environments with the Franka Emika Robot (Panda). They range from easy tasks like a lifting a cube to sequential multi-step tasks like picking up a peg and inserting it into a hole.}
\vspace{-0.4cm}
\label{fig:scenarios}
\end{figure}

\par In this work we present Behavior-based Bayesian Optimization and Planning (\ours)~that generates BTs by building a reactive tree structure using a planner and then subsequently learns the BT parameters with Bayesian Optimization (BO). With the tree structure as a prior, BO can then focus on tuning parameters that are difficult to plan and reason about. The method is evaluated in a simulation environment with eight different manipulation tasks. 

It drastically outperforms the award-winning state-of-the-art RL algorithm \textit{MAPLE}~\cite{nasiriany2022augmenting} in terms of the number of simulation steps needed to learn to solve the tasks while using exactly the same behavior primitives. By induction this also means that \ours~is much more efficient than popular algorithms like \textit{HIRO}~\cite{nachum2018data} and \textit{DAC}~\cite{zhang2019dac}. The speedup
could even enable training on real robot systems instead of just simulation as a lot fewer evaluations are needed. Furthermore, it is also shown that our method is less dependent on reward shaping in the form of affordances compared to the benchmark method.

Lastly, another advantage of the modular tree structure is that it also allows the task to be divided into subtasks and learned in sequence for even faster progress.

The main contributions of this paper are:
\begin{itemize}
    \item A novel method, \ours{}, that combines reactive planning with efficient parameter tuning to yield state-of-the-art learning performance, while leading to an interpretable and robust policy.
    \item A new method to calculate the uncertainty of a random forest surrogate model within BO that outperforms the standard method on several tasks. 
    \item A set of experiments verifying and validating our approach in comparison to the state-of-the-art RL method \emph{MAPLE}\cite{nasiriany2022augmenting}, showing that \ours~learns to solve the given tasks up to 46 times faster.
 
\end{itemize}

\section{Background and Related Work}
In this section we provide the relevant background on behavior trees and Bayesian optimization and discuss the related work.

\begin{figure}
    \setlength{\fboxrule}{0pt}
		\framebox{\parbox{3in}{
            \centering

\tabskip=0pt
\valign{#\cr
  \noalign{\hfill}
  \hbox{%
    \begin{subfigure}{.24\textwidth}
    \centering
    \includegraphics[width=0.4\textwidth]{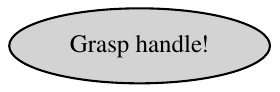}
    \caption{Subtree for $n=1$}
    \end{subfigure}%
  }\vfill
  \hbox{%
    \begin{subfigure}{.24\textwidth}
    \centering
    \includegraphics[width=\textwidth]{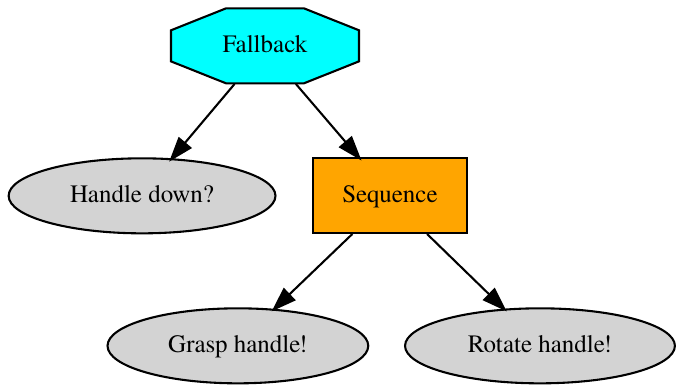}
    \caption{Subtree $n=2$}
    \end{subfigure}%
  }\cr
  \hbox{%
    \begin{subfigure}[b]{.24\textwidth}
    \centering
    \includegraphics[width=\textwidth]{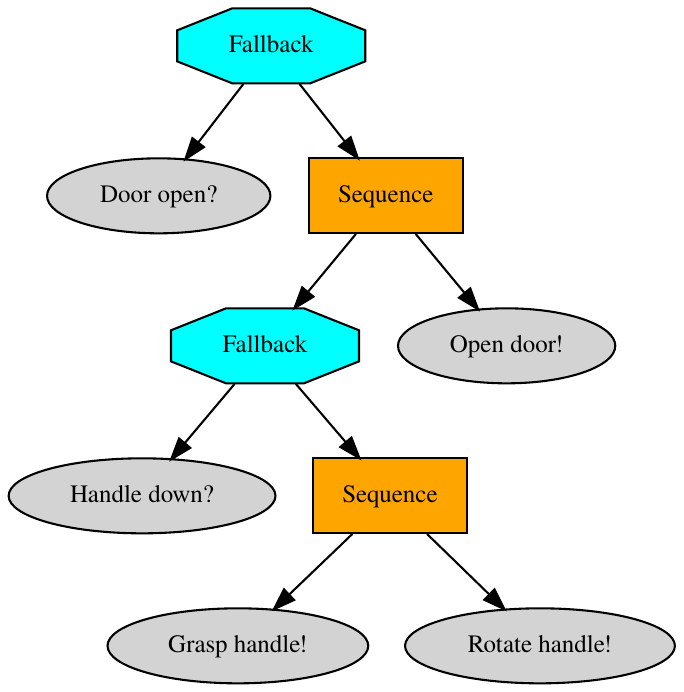}
    \caption{Full tree}
    \end{subfigure}%
  }\cr
}
}
}
\caption{Different subtrees for a door opening scenario. In a cascaded learning setup, the method starts by learning how to grasp the handle (a) before the tree gets extended by the handle rotation (b) and finally the full tree is constructed (c).}
\label{fig:bt}
\vspace{-0.4cm}
\end{figure}

\subsection{Behavior Trees}
Behavior Trees (BTs) were first used in the computer game industry, but have recently seen increased use in robotics~\cite{colledanchise_behavior_2018, iovino_survey_2022}.
A BT is a directed tree where a tick signal propagates from the root node down to the leaves. The nodes are executed only when they receive the tick signal and return one of the states \textit{Success}, \textit{Failure} and \textit{Running}. The non-leaf nodes are called \emph{control flow nodes}. The flow nodes most commonly used are \emph{Sequence}, which ticks children sequentially from left to right, returning once all succeed or one fails, and \emph{Fallback} (or \emph{Selector}) which also runs sequentially but returns when one succeeds or all fail. Leaves are called \emph{execution nodes} or \emph{behaviors} and are usually separated into the types \textit{Action}("!") and \textit{Condition}("?"). Conditions encode status checks and sensory readings, only returning \textit{Success} or \textit{Failure} while actions encode robot skills that can take more than one tick to complete and therefore can also return \textit{Running}. Figure~\ref{fig:bt} shows three example BTs for a door opening scenario.

The main advantages of BTs are that they are readable and have inherent support for task hierarchy, action sequencing and reactivity. They are also inherently modular~\cite{colledanchise_behavior_2018}, in fact even optimally modular~\cite{biggar2022modularity}. The \emph{Running} return state grants the reactivity property because a running action can be preempted by higher priority ones. BTs have been shown to improve on other representations, such as finite state machines, especially in terms of modularity and reactivity~\cite{iovino_programming_2022,colledanchise_how_2017,biggar_modularity_2022}.

\subsection{Bayesian Optimization}
In many practical optimization problems, there is no closed form expression available for the function to optimize. Instead, the user can only interact with the system by first selecting a configuration to evaluate and subsequently observing its performance. Often this black-box function is also expensive to evaluate; in the particular setting used in this paper, it amounts to running a simulation of a robot performing the specified task. 

Bayesian optimization (BO) is a paradigm developed to efficiently optimize such problems, while limiting the number evaluations. It has recently shown great performance in a variety of applications, such as robotics \cite{ calandra2016bayesian, mayr2022skill, rai2018bayesian}, hyperparameter tuning \cite{klein2017fast, kandasamy2018neural, ru2020interpretable}, and material design \cite{frazier2015bayesian, packwood2017bayesian, hughes2021tuning}.

We consider the problem of finding a global maximum of an unknown black-box objective function $f$: $\mathbf{s^*} \in \mathop{\mathrm{arg\,max}}_{\mathbf{s} \in \mathbb{S}}
f(\mathbf{s}),$ over some pre-specified domain $\mathbb{S}$ in $D$ dimensions. The variables defining $\mathbb{S}$ can be real, integer, ordinal and categorical~\cite{nardi18hypermapper}. We further assume that the evaluations of $f$ are disturbed by observation noise and do not provide information on the function gradients.

BO is a sequential approach that iteratively selects new configurations to evaluate, trading off exploration and exploitation. It uses a surrogate model of the objective function and effectively learns the function as it gathers more data. The most common models are Gaussian Processes (GPs) \cite{williams2006gaussian} for their natural ability to quantify uncertainty on top of yielding accurate predictions and Random Forests (RFs) \cite{lindauer2022smac3, shahriari2015taking} for their versatility and scalability to a higher number of samples. Which configuration to select next is chosen by maximizing an acquisition function that quantifies the exploration-exploitation trade-off. Common examples are the \textit{expected improvement} or \textit{upper confidence bound}.
For a more thorough introduction to BO, see \cite{frazier2018tutorial}.

\subsection{Related work}
Various combinations of planning and RL have previously been proposed for other domains in~\cite{grounds_combining_2008,faust2018prm,francois-lavet_combined_2019,moerland2023model}. In particular, \cite{styrud_combining_2022} uses BTs as the underlying structure, and \cite{koza_genetic_1992,sloss_2019_2020} combine a planner with genetic programming. Using genetic programming to learn BTs has been done primarily for computer games~\cite{colledanchise_learning_2019}, but there are also examples for robotic manipulation applications~\cite{iovino2021learning, styrud_combining_2022,iovino2023framework}. A more extensive analysis is given in~\cite{iovino_survey_2022}.
The combination of a sequential planner and learning with BTs was also proposed in~\cite{mayr2022combining, mayr2022skill} for a peg-insertion and a pushing task. In~\cite{mayr22priors} it is shown how priors defined by operators or based on experience can accelerate learning and increase the safety during learning. As an extension, \cite{ahmad2023learning} learns a GP model to generalize to task variations.
\par 

Regarding automated planners, we use an adaptation of the Planning Domain Definition Language (PDDL)-style planner from~\cite{colledanchise_towards_2019} that creates BTs by leveraging backchaining. We build on the later adaptations of the same planner from~\cite{styrud_combining_2022,gustavsson_combining_2022, iovino2023framework}. The main advantage of this planner is its simplicity, but there are other more advanced planners for BTs as well. Some examples are Linear Temporal Logic (LTL)~\cite{tumova_maximally_2014,colledanchise_synthesis_2017} 
and Hierarchical-Task-Network (HTN) planning~\cite{holzl_reasoning_2015, rovida_extended_2017}. There are also other PDDL-style planners and we refer to Section~4.2 in~\cite{iovino_survey_2022} for a more exhaustive list.

We compare our method against RL with parameterized actions~\cite{masson2016reinforcement, dalal2021accelerating} - specifically with \emph{MAPLE}~\cite{nasiriany2022augmenting}. Although these are completely different algorithms from ours, the types of problems they solve are essentially the same.

\section{Approach}
\label{sec:approach}
The key assumption in this work is that if there exists a set of parameterized actions that a robot can execute, these actions were likely designed with an intended use and effect on the robots environment. It is then possible to create a plan by using the actions under the assumption that for some values of the action's parameters, the actions will succeed and work in the intended way. Utilizing this, our proposed method consists of using a planner to obtain the structure of the BT and then employing a BO algorithm in an RL framework to tune the parameters of the nodes of the BT. The complete code of the planner and all other algorithms are available online\footnote{\url{https://github.com/jstyrud/BeBOP}}.
\subsection{Planner}
In this paper we use a PDDL planner adapted from~\cite{colledanchise_towards_2019} that was later extended in~\cite{styrud_combining_2022, gustavsson_combining_2022, iovino2023framework}.
As input to the planner, all behaviors have a set of preconditions that must be fulfilled in order to execute the behavior successfully and a set of postconditions that can be expected to be fulfilled when the behavior is done. A set of goal conditions defines the robot's task. Starting with goal conditions and proceeding backwards, the actions that complete the task or fulfill the necessary conditions for other actions are found iteratively and expanded until all conditions have been met. In this work, we improve the planner from~\cite{gustavsson_combining_2022} with some additions. Mainly, we trim the resulting tree by \textit{a)} removing any control nodes with only a single child and by \textit{b)} removing any post-condition nodes that are placed directly to before the corresponding action. The latter step assumes that all behaviors check for post-conditions internally before executing. We further introduce the concept of composite subtrees where, after planning, certain leaves can either be expanded into subtrees with multiple nodes or they can be replaced by some other parameterized behavior. This allows us to exploit the fact that, e.g., the behaviors \emph{Reach} and \emph{Open} together comprise a \emph{Place} skill and that a \emph{Reach} can generalize different behaviors at planning time such as opening a door or moving a grasped object.

\subsection{Optimization}
\label{sec:optimization}
In order to optimize a policy, we follow the policy-search formulation~\cite{deisenroth13r,chatzilygeroudis2019survey,chatzilygeroudis172iicirsi}. The goal is to find a policy $\pi, \mathbf{u}~=~\pi(\mathbf{x}|\boldsymbol{\theta})$ with policy parameters $\boldsymbol{\theta}$ such that we maximize the expected long-term reward when executing the policy for $T$ time steps.
Here, we use BO to tune the parameters. A given (planned) BT has a set of action nodes, such that each node can have zero or more parameters that can be learned for a given task. To construct a learning problem, we automatically examine a BT to obtain these adjustable parameters, their domains and dependencies. 

We use a customized version of the BO implementation in \textit{hypermapper}~\cite{nardi18hypermapper} as it supports a wide range of variable types and user priors for the optimum \cite{hvarfner2022pi}.

To provide a robust reward measure for the surrogate model and to prevent BO from overfitting to the training data, we evaluate each set of parameters in up to 20 episodes using \emph{robotsuite's} domain randomization of the tasks with different seeds.
We initially run each set of parameters for three episodes in different randomizations of the task simulation. After that, we estimate the variance after each episode and calculate the probability that this policy will outperform the current best policy. We then continue with more episodes as long as the probability is above 5 percent or until we have reached 20 episodes. We evaluate the parameter sets on the same random seeds for the training episodes to increase consistency of the evaluation results when training the surrogate model.

\subsection{Improved Random Forest Surrogate}

While GPs are the most common choice of surrogate models in BO, they assume a certain level of smoothness of the objective function, that is often not satisfied in the robotics tasks we consider. For example when inserting a peg into a hole, a millimeter change in offset in a movement primitive can result in a drastic difference in reward. Because of this, we instead use random forests which are more amenable to non-smooth objective functions. However, in contrast to GPs, RFs do not innately provide a variance estimate for its predictions, which is an essential building block in the BO selection process. Hutter et al. proposed the use of the empirical prediction variance across trees \cite{hutter2011sequential}, but this suffers from that the uncertainty does not inherently grow further away from previously observed data. This, in turn, hampers exploration. 

\begin{figure}[tpb]
    \setlength{\fboxrule}{0pt}
		\framebox{\parbox{3in}{
            \centering
            \includegraphics[width=\columnwidth]{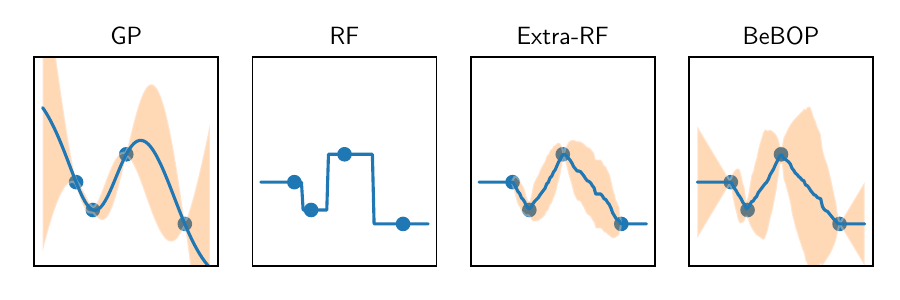}
        }
    }
    \vspace{-0.3cm}
    \caption{The predicted mean and standard deviation on a toy example for 4 different models: Gaussian Processes (GP), Random Forests (RFs), Extremely randomized RFs and our proposed uncertainty measure with an additional linear standard deviation term.}
    \vspace{-0.4cm}
    \label{fig:models}
\end{figure}

To improve the performance of the optimization, we propose two adjustments to the RF model. First, we use extremely randomized trees which randomize among all optimal splits in  each tree \cite{geurts2006extremely}. This makes the prediction surface much smoother, which makes for a richer predicted function surface. This was previously used by, for example, \cite{nardi18hypermapper} and \cite{wu2022autotuning}. Secondly, we propose a new uncertainty metric, that extends the standard deviation estimate proposed by Hutter et al. by a term proportional to the distance to the closest previous observation. This incentivizes the optimization to continue exploring. To the best of our knowledge, this is the first paper to suggest such a modification in a BO setting. In Fig.~\ref{fig:models}, we show the impact of the modifications.
As we will see in Section \ref{sec:results}, this significantly improves the results.

\subsection{Combining Planning and Bayesian Optimization}
To combine the planner with the BO algorithm, we first run the planner on each task to obtain the BT structure. The planner, however, leaves a number of parameters unsolved for each behavior, and during planning it assumes that there exists some set of parameters for which the behaviors will succeed. These parameters are then given as input to BO. For each parameter, the behaviors also specify an upper and lower bound and these limits should be grounded in the actual application.

\textbf{Cascaded Learning:} We also note that because of the hierarchical nature of the BTs that are output from the planner, the BTs can be divided into a number of subtrees that can be run sequentially with gradually larger subtrees, representing subtasks. For example, grasping would be a sub-task of moving an object. In this way, we can learn the smaller subtask first, using parameters from the solution as priors for the optimum to the next, larger subtree. This way we can potentially speed up the optimization as the learning time typically scales super-linearly with the number of parameters. We call this version of our method \emph{cascaded \ours}.
Starting with $n=1$, to find subtree $n$ we start with the first action node and traverse the tree left to right, depth first, counting the action nodes. Continue up to but not including the action node $n+1$. The last node before the action node $n+1$ will be the last node in the subtree. Action nodes without free parameters are omitted, as they do not increase the complexity of our optimization problem. All subtrees for smaller $n$ will also be included in subtree $n$. Fig.~\ref{fig:bt} shows examples of the resulting subtrees for a door opening scenario. Note that the subtree for $n=1$ consists of only one node and that the behaviors in the figure are only aliases during planning of more generic behaviors, as listed in Section~\ref{sec:behaviors}.
For every $n$-th subtree, we run the BO in batches of 50 iterations until no improvement is seen since the last batch. We then use the best solution found as priors and run BO on the subtree $n+1$.
Splitting learning tasks into subtrees has also been proposed in~\cite{mayr2021learning}, but a) without an automated procedure to do this and b) while keeping the previously learned parameters fixed when combining trees.
The advantage we get by \emph{not} fixing the parameters of previous subtrees is that the optimal values might be different in the context of the complete tree. 

\section{Experimental Setup}
\label{sec:exper}
We benchmark our algorithm using \textit{robosuite}~\cite{zhu2020robosuite} in the eight simulated robot manipulation scenarios shown in Fig.~\ref{fig:scenarios} that are used in \emph{MAPLE}~\cite{nasiriany2022augmenting}, the method that we compare against. One feature of \emph{robosuite} is that it supports slight domain randomizations of the task environments to allow for the evaluation of the robustness of policies. In the \emph{Door} scenario in the original benchmark the robot was incapable of grasping the door handle in all instances of domain randomization. Therefore we made a small change by allowing the gripper to rotate in the same way as in the other tasks. 
This change had no noticeable impact on the performance of \emph{MAPLE} on the task.

In this paper, we refrained from performing additional validation of the resulting policies learned in simulation on real robot systems. It has been shown previously, including in the referred \emph{MAPLE} paper which uses the same behaviors~\cite{nasiriany2022augmenting}, that policies acting at this abstraction level are easily transferable from simulation~\cite{styrud_combining_2022, mayr2021learning, mayr2022combining, mayr2022skill, mayr22priors, ahmad2023learning}.

\begin{figure*}[t!!]
    \setlength{\fboxrule}{0pt}
		\framebox{\parbox{3in}{
            \centering
                \includegraphics[width=1.0\textwidth]{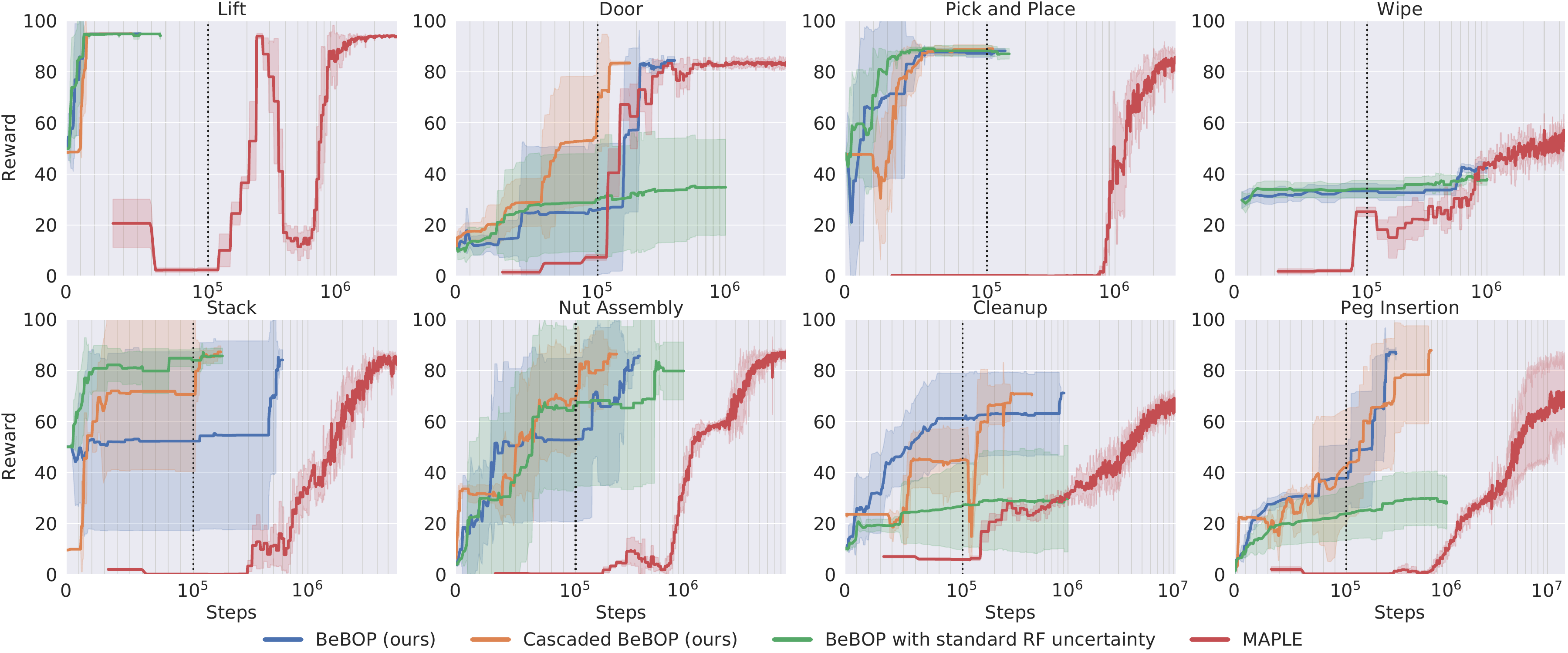}
            }
        }
\caption{Episodic reward learning curves for the eight tasks. The x-axis is linear until 10\textsuperscript{5} steps and logarithmic thereafter as marked by the dotted vertical line. The results highlight the importance of our new uncertainty measure and that our approaches outperform \textit{MAPLE} by a large margin.}
\vspace{-0.4cm}
\label{fig:fit_learning_curve}
\end{figure*}

\subsection{Behaviors}
\label{sec:behaviors}
We use the same action primitives as \emph{MAPLE} and only communicate with the simulation using the same action and observation vectors as the neural networks in the original benchmarks.
We wrap the input and output vectors of the simulation with behaviors that can be used by a behavior tree. The five behaviors correspond to the five primitives used by \textit{MAPLE} and call the corresponding parameterized behavior primitives when executing:
\begin{itemize}
    \item \textbf{Reach:} The robot moves to some position relative to an object or the origin of the coordinate frame. The offset (x, y, z) is specified by the behavior parameters.
    \item \textbf{Grasp:} The robot moves to some position relative to a graspable object and closes the gripper. The offset (x, y, z) and yaw angle are the behavior parameters.
    \item \textbf{Push:} The robot attempts to push an object towards an x, y position by moving to the opposing side of the object and pushing toward the goal position. The behavior parameters are the coordinates x and y.
    \item \textbf{Open:} The robot opens the gripper. No parameters.
    \item \textbf{Atomic:} The robot applies an atomic action for one step, moving a delta calculated from the relative distance between some object position and some (x, y, z) and yaw angle as specified by the behavior parameters.
\end{itemize}
In addition to behaviors, the BT also needs conditions that process the information in the observation vector. In the experiments, we make use of three different condition nodes.
\begin{itemize}
    \item \textbf{At:} Returns \textit{Success} if the object is at a given position. This is exclusively used for goal conditions and the parameters are therefore fixed.
    \item \textbf{Angle >:} Checks if the angle of some object is larger than a value parameter.
    \item \textbf{Aligned:} Checks the observation vector to see if the object is aligned. No parameters except the object identifier. Used only in the peg-insertion task.
\end{itemize}
We use pre- and post-conditions of the behaviors for planning and because the behaviors work with relative coordinates, we claim that in most real robot applications and frameworks, such as~\cite{mayr23iros}, this knowledge will be readily available.
The BTs are implemented using the \emph{PyTrees} framework\footnote{\url{https://github.com/splintered-reality/py_trees}, version 2.2.2. Specifically, we use a forked version with slightly changed visuals: \url{https://github.com/jstyrud/py_trees}}.

\subsection{Reward}
\label{sec:reward}
The reward for the episodes is the same dense rewards with affordances as the original benchmark~\cite{nasiriany2022augmenting}. However, since BTs have a natural stop condition when the tree returns \textit{Success} or \textit{Failure}, we can also handle that. We want to avoid solutions where the tree returns \textit{Failure}, so we add a negative reward of $-500.0$ for any such episode. If the BT stops before the maximum number of steps in the environment, we assume that the same reward that was given for the last step would be given for the rest of the episode and extrapolate it until the maximum number of steps.
As in~\cite{nasiriany2022augmenting} we evaluate the policy in 20 validation episodes with different random seeds than during training. We only validate the currently best BT found, based on the reward of the training episodes. The affordances and failure penalty are not used for the validation.
The affordance penalty as described in~\cite{nasiriany2022augmenting} is a form of reward shaping where a penalty is given for an action that is performed outside a manually specified region. In~\cite{nasiriany2022augmenting} it is shown that \emph{MAPLE} is unable to make progress even on the simple \emph{Pick and Place} task without the affordances.

\section{Results}
\label{sec:results}

Figure~\ref{fig:fit_learning_curve} shows learning curves for all eight tasks as the mean of five separate repetitions (same as in \cite{nasiriany2022augmenting}) for each method with shaded areas denoting the standard deviation. We run the experiments until the task is solved, but maximally 10\textsuperscript{6} time steps. The figure shows that our method outperforms \emph{MAPLE} in 7 of the 8 benchmarks. The number of steps needed to solve a task is often an order of magnitude less than for \emph{MAPLE}. By outperforming \emph{MAPLE}~\cite{nasiriany2022augmenting} we consequently also outperform other RL algorithms like \textit{HIRO}~\cite{nachum2018data} and \textit{DAC}~\cite{zhang2019dac} which are not shown here but are discussed in detail in~\cite{nasiriany2022augmenting}. 
We also note that our new uncertainty measure as described in Section \ref{sec:optimization} performs equally or better than the standard RF uncertainty measure (green) for all benchmark tasks except \emph{Stack}. We believe that this is simply because \emph{Stack} is easy enough to solve without it. However, BO with the standard measure fails to solve the more difficult tasks even with more iterations.

The cascaded version of \ours{} (orange) performs even better on several benchmarks, especially those that can be logically divided into a sequence of subtasks.


For the \emph{Wipe} task none of the methods, including \emph{MAPLE}, are able to solve the task. It is likely that the allowed time is in fact insufficient to perform the task reliably. The markers to wipe are randomly placed, and it is possible that MAPLE can statistically learn their positions, giving it a slight edge in this benchmark. However, with the only observation of the markers being their average position, not enough information is given to make an efficient wiping pattern.


\begin{figure}[tpb]
    \setlength{\fboxrule}{0pt}
		\framebox{\parbox{3in}{
            \centering
            \includegraphics[width=0.48\textwidth, left]{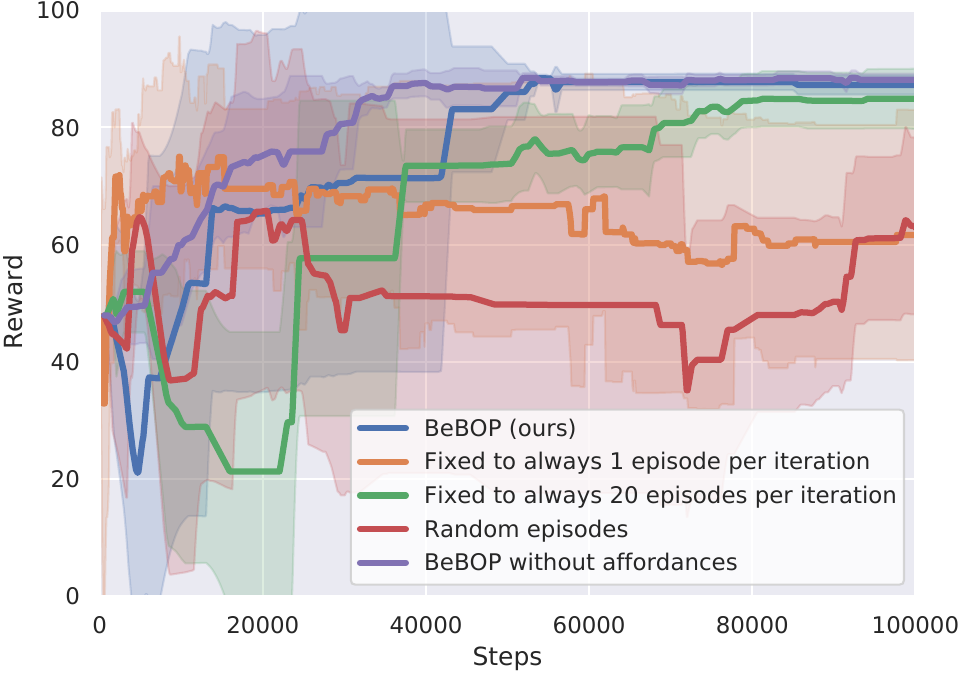}
            }
        }
\caption{The reward of the best policy on the validation data for the \emph{Pick and Place task}. It shows that our evaluation strategy (blue) does not only learn the fastest and most robust, but our method can also learn without the special affordances (purple).}
\vspace{-0.4cm}
\label{fig:ablations}
\end{figure}

We also studied how the evaluation procedure impacts \ours's performance. In Fig.~\ref{fig:ablations} we see that solving the \emph{Pick and Place} task without affordances (purple) has no significant impact on the learning rate, while \emph{MAPLE}~\cite{nasiriany2022augmenting} was reported to make no progress at all without affordances.
In the same figure, we show learning curves for different variants of choosing the training episodes per parameter set. Fixing the number of episodes to 1 (orange) quickly finds a good solution on the training data, but fails to generalize to the validation episodes. Fixing it to always run 20 (green) will generalize to the validation episodes but wastes a lot of steps on poor solutions which results in slow learning. Choosing 20 random environments for evaluations can mean that a policy gets a set of easy or hard environments to evaluate in, which makes a comparison difficult. Such inconsistencies also makes it harder to fit the BO surrogate model. The red line in Fig.~\ref{fig:ablations} shows how this negatively affects the learning performance. Finally, our proposed iterative evaluation shown in blue learns fast and achieves robust results.

\begin{figure}[tpb]
    \setlength{\fboxrule}{0pt}
		\framebox{\parbox{3in}{
            \centering
                \includegraphics[width=0.48\textwidth]{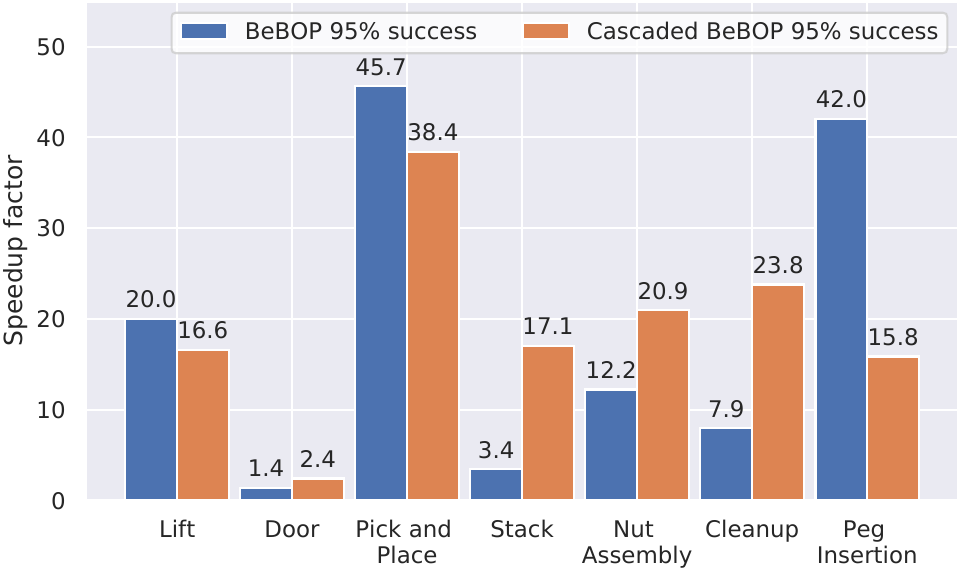}
            }
        }
\caption{The speedup factor when compared to \emph{MAPLE} on the 7 tasks that are solved. The factor is computed at reaching a mean 95\% success rate in the validation data.}
\vspace{-0.4cm}
\label{fig:barplot}
\end{figure}

In Fig.~\ref{fig:barplot} we show the actual speedup factor to reach the specified task success rates for \ours~and \textit{Cascaded \ours} compared to \emph{MAPLE} for the 7 tasks that were solved. The plot compares the point where the mean success rate of the five repetitions has reached $95\%$ for the first time. In our experiments \emph{MAPLE} failed to solve \emph{Peg Insertion} task in one of the repetitions, so we only use the other four.
As shown in the figure, even the smallest speedup of \ours~compared to \textit{MAPLE} in the \textit{Door} task is still 1.4. For many tasks, \ours~learns more than 15 times faster and reaches a speedup of up to $46$ times. This not only saves a lot of compute time when learning in simulation, but also makes learning on a real robotic system much more realistic.

\section{Conclusions}
We present \ours, a method for combining reactive task planning and Bayesian optimization to create behavior-tree policies. We show that our method for learning these policies outperforms state-of-the-art RL algorithms such as \textit{MAPLE, HICO} and \textit{DAC} by a large margin. While using exactly the same behavior primitives, our method solves the robotic manipulation benchmarks using on average only about $5\%$ of the steps needed by \textit{MAPLE}.
Those results are further improved by utilizing the structure of the behaviour trees to divide the task into a sequence of sub-tasks, which makes \ours{} solve many of the benchmarks even faster. 
An ablation analysis indicates that \ours~is also less dependent on reward shaping in the form of special affordances compared to the benchmark method. Our newly introduced uncertainty measure for the random forest surrogate model accelerates and robustifies the learning with Bayesian optimization in robotics tasks significantly.

At the same time, the obtained BT policies are deterministic as well as interpretable and modifyable by humans. This makes them much more attractive for sensitive environments such as in industrial manufacturing or private households.

\section{Future work}
One natural direction for future work is to use a more advanced planner and to combine it with platforms that support reasoning such as~\cite{mayr23iros, mayr2023using}, as both leave less for the optimization algorithm to learn.
In addition, it would be interesting to study the possibility of re-planning the BT structure in case of changes in the environment without having to re-learn all parameters from scratch.

We believe that our method can generalize to many other types of task, perhaps also outside the robotic domain and confirming that would be an intuitive next step.

Certain tasks are well suited for neural networks, such as when decisions need to factor in many different variables. In those cases, it could be a good approach to to include a network as part of a BT policy. That way, the advantages of a transparent and modular BT are kept for the majority of the policy, while still enjoying the strength of the neural network~\cite{sprague2022adding}.


\clearpage
\bibliographystyle{IEEEtran}
\bibliography{IEEEabrv,biblio, references}

\begin{thebibliography}{10}
\providecommand{\url}[1]{#1}
\csname url@rmstyle\endcsname
\providecommand{\newblock}{\relax}
\providecommand{\bibinfo}[2]{#2}
\providecommand\BIBentrySTDinterwordspacing{\spaceskip=0pt\relax}
\providecommand\BIBentryALTinterwordstretchfactor{4}
\providecommand\BIBentryALTinterwordspacing{\spaceskip=\fontdimen2\font plus
\BIBentryALTinterwordstretchfactor\fontdimen3\font minus
  \fontdimen4\font\relax}
\providecommand\BIBforeignlanguage[2]{{%
\expandafter\ifx\csname l@#1\endcsname\relax
\typeout{** WARNING: IEEEtran.bst: No hyphenation pattern has been}%
\typeout{** loaded for the language `#1'. Using the pattern for}%
\typeout{** the default language instead.}%
\else
\language=\csname l@#1\endcsname
\fi
#2}}

\bibitem{nasiriany2022augmenting}
S.~Nasiriany, H.~Liu, and Y.~Zhu, ``Augmenting reinforcement learning with
  behavior primitives for diverse manipulation tasks,'' in \emph{2022
  International Conference on Robotics and Automation (ICRA)}.\hskip 1em plus
  0.5em minus 0.4em\relax IEEE, 2022, pp. 7477--7484.

\bibitem{ghallab_automated_2016}
M.~Ghallab, D.~Nau, and P.~Traverso, \emph{\BIBforeignlanguage{en}{Automated
  {Planning} and {Acting}}}.\hskip 1em plus 0.5em minus 0.4em\relax Cambridge
  University Press, Aug. 2016.

\bibitem{iovino_survey_2022}
M.~Iovino, E.~Scukins, J.~Styrud, P.~{\"O}gren, and C.~Smith, ``A survey of
  {{Behavior Trees}} in robotics and {{AI}},'' \emph{Robotics and Autonomous
  Systems}, vol. 154, p. 104096, Aug. 2022.

\bibitem{colledanchise_behavior_2018}
M.~Colledanchise and P.~{\"O}gren, \emph{Behavior {{Trees}} in {{Robotics}} and
  {{AI}} : {{An Introduction}}}.\hskip 1em plus 0.5em minus 0.4em\relax {CRC
  Press}, July 2018.

\bibitem{colledanchise_how_2017}
------, ``How {{Behavior Trees Modularize Hybrid Control Systems}} and
  {{Generalize Sequential Behavior Compositions}}, the {{Subsumption
  Architecture}}, and {{Decision Trees}},'' \emph{IEEE Transactions on
  Robotics}, vol.~33, no.~2, pp. 372--389, Apr. 2017.

\bibitem{nachum2018data}
O.~Nachum, S.~S. Gu, H.~Lee, and S.~Levine, ``Data-efficient hierarchical
  reinforcement learning,'' \emph{Advances in neural information processing
  systems}, vol.~31, 2018.

\bibitem{zhang2019dac}
S.~Zhang and S.~Whiteson, ``Dac: The double actor-critic architecture for
  learning options,'' \emph{Advances in Neural Information Processing Systems},
  vol.~32, 2019.

\bibitem{biggar2022modularity}
O.~Biggar, M.~Zamani, and I.~Shames, ``On modularity in reactive control
  architectures, with an application to formal verification,'' \emph{ACM
  Transactions on Cyber-Physical Systems (TCPS)}, vol.~6, no.~2, pp. 1--36,
  2022.

\bibitem{iovino_programming_2022}
M.~Iovino, J.~F{\"o}rster, P.~Falco, J.~J. Chung, R.~Siegwart, and C.~Smith,
  ``On the programming effort required to generate {{Behavior Trees}} and
  {{Finite State Machines}} for robotic applications,'' in \emph{2023 {{IEEE
  International Conference}} on {{Robotics}} and {{Automation}} ({{ICRA}})},
  May 2023.

\bibitem{biggar_modularity_2022}
O.~Biggar, M.~Zamani, and I.~Shames, ``On {{Modularity}} in {{Reactive Control
  Architectures}}, with an {{Application}} to {{Formal Verification}},''
  \emph{ACM Transactions on Cyber-Physical Systems}, vol.~6, no.~2, pp.
  19:1--19:36, Apr. 2022.

\bibitem{calandra2016bayesian}
R.~Calandra, A.~Seyfarth, J.~Peters, and M.~P. Deisenroth, ``Bayesian
  optimization for learning gaits under uncertainty: An experimental comparison
  on a dynamic bipedal walker,'' \emph{Annals of Mathematics and Artificial
  Intelligence}, vol.~76, pp. 5--23, 2016.

\bibitem{mayr2022skill}
M.~Mayr, F.~Ahmad, K.~Chatzilygeroudis, L.~Nardi, and V.~Krueger, ``Skill-based
  multi-objective reinforcement learning of industrial robot tasks with
  planning and knowledge integration,'' in \emph{2022 IEEE International
  Conference on Robotics and Biomimetics (ROBIO)}.\hskip 1em plus 0.5em minus
  0.4em\relax IEEE, 2022, pp. 1995--2002.

\bibitem{rai2018bayesian}
A.~Rai, R.~Antonova, S.~Song, W.~Martin, H.~Geyer, and C.~Atkeson, ``Bayesian
  optimization using domain knowledge on the atrias biped,'' in \emph{2018 IEEE
  International Conference on Robotics and Automation (ICRA)}.\hskip 1em plus
  0.5em minus 0.4em\relax IEEE, 2018, pp. 1771--1778.

\bibitem{klein2017fast}
A.~Klein, S.~Falkner, S.~Bartels, P.~Hennig, and F.~Hutter, ``Fast bayesian
  optimization of machine learning hyperparameters on large datasets,'' in
  \emph{Artificial intelligence and statistics}.\hskip 1em plus 0.5em minus
  0.4em\relax PMLR, 2017, pp. 528--536.

\bibitem{kandasamy2018neural}
K.~Kandasamy, W.~Neiswanger, J.~Schneider, B.~Poczos, and E.~P. Xing, ``Neural
  architecture search with bayesian optimisation and optimal transport,''
  \emph{Advances in neural information processing systems}, vol.~31, 2018.

\bibitem{ru2020interpretable}
B.~Ru, X.~Wan, X.~Dong, and M.~Osborne, ``Interpretable neural architecture
  search via bayesian optimisation with weisfeiler-lehman kernels,''
  \emph{arXiv preprint arXiv:2006.07556}, 2020.

\bibitem{frazier2015bayesian}
P.~I. Frazier and J.~Wang, ``Bayesian optimization for materials design,'' in
  \emph{Information science for materials discovery and design}.\hskip 1em plus
  0.5em minus 0.4em\relax Springer, 2015, pp. 45--75.

\bibitem{packwood2017bayesian}
D.~Packwood \emph{et~al.}, \emph{Bayesian optimization for materials
  science}.\hskip 1em plus 0.5em minus 0.4em\relax Springer, 2017.

\bibitem{hughes2021tuning}
Z.~E. Hughes, \emph{et~al.}, ``Tuning materials-binding peptide sequences
  toward gold-and silver-binding selectivity with bayesian optimization,''
  \emph{ACS nano}, vol.~15, no.~11, pp. 18\,260--18\,269, 2021.

\bibitem{nardi18hypermapper}
L.~Nardi, D.~Koeplinger, and K.~Olukotun, ``Practical design space
  exploration,'' in \emph{International Symposium on Modeling, Analysis, and
  Simulation of Computer and Telecommunication Systems}, 2019.

\bibitem{williams2006gaussian}
C.~K. Williams and C.~E. Rasmussen, \emph{Gaussian processes for machine
  learning}.\hskip 1em plus 0.5em minus 0.4em\relax MIT press Cambridge, MA,
  2006, vol.~2, no.~3.

\bibitem{lindauer2022smac3}
M.~Lindauer, \emph{et~al.}, ``Smac3: A versatile bayesian optimization package
  for hyperparameter optimization,'' 2022.

\bibitem{shahriari2015taking}
B.~Shahriari, K.~Swersky, Z.~Wang, R.~P. Adams, and N.~De~Freitas, ``Taking the
  human out of the loop: A review of bayesian optimization,'' \emph{Proceedings
  of the IEEE}, vol. 104, no.~1, pp. 148--175, 2015.

\bibitem{frazier2018tutorial}
P.~I. Frazier, ``A tutorial on bayesian optimization,'' \emph{arXiv preprint
  arXiv:1807.02811}, 2018.

\bibitem{grounds_combining_2008}
M.~Grounds and D.~Kudenko, ``\BIBforeignlanguage{en}{Combining {Reinforcement}
  {Learning} with {Symbolic} {Planning}},'' in
  \emph{\BIBforeignlanguage{en}{Adaptive {Agents} and {Multi}-{Agent} {Systems}
  {III}. {Adaptation} and {Multi}-{Agent} {Learning}}}, ser. Lecture {Notes} in
  {Computer} {Science}, K.~Tuyls, A.~Nowe, Z.~Guessoum, and D.~Kudenko,
  Eds.\hskip 1em plus 0.5em minus 0.4em\relax Berlin, Heidelberg: Springer,
  2008, pp. 75--86.

\bibitem{faust2018prm}
A.~Faust, \emph{et~al.}, ``Prm-rl: Long-range robotic navigation tasks by
  combining reinforcement learning and sampling-based planning,'' in \emph{2018
  IEEE international conference on robotics and automation (ICRA)}.\hskip 1em
  plus 0.5em minus 0.4em\relax IEEE, 2018, pp. 5113--5120.

\bibitem{francois-lavet_combined_2019}
V.~Francois-Lavet, Y.~Bengio, D.~Precup, and J.~Pineau,
  ``\BIBforeignlanguage{en}{Combined {Reinforcement} {Learning} via {Abstract}
  {Representations}},'' \emph{\BIBforeignlanguage{en}{Proceedings of the AAAI
  Conference on Artificial Intelligence}}, vol.~33, no.~01, pp. 3582--3589,
  July 2019.

\bibitem{moerland2023model}
T.~M. Moerland, J.~Broekens, A.~Plaat, C.~M. Jonker, \emph{et~al.},
  ``Model-based reinforcement learning: A survey,'' \emph{Foundations and
  Trends{\textregistered} in Machine Learning}, vol.~16, no.~1, pp. 1--118,
  2023.

\bibitem{styrud_combining_2022}
J.~Styrud, M.~Iovino, M.~Norrl{\"o}f, M.~Bj{\"o}rkman, and C.~Smith,
  ``Combining {{Planning}} and {{Learning}} of {{Behavior Trees}} for {{Robotic
  Assembly}},'' in \emph{2022 {{International Conference}} on {{Robotics}} and
  {{Automation}} ({{ICRA}})}, May 2022, pp. 11\,511--11\,517.

\bibitem{koza_genetic_1992}
J.~R. Koza, \emph{\BIBforeignlanguage{en}{Genetic {Programming}: {On} the
  {Programming} of {Computers} by {Means} of {Natural} {Selection}}}.\hskip 1em
  plus 0.5em minus 0.4em\relax MIT Press, 1992.

\bibitem{sloss_2019_2020}
A.~N. Sloss and S.~Gustafson, ``\BIBforeignlanguage{en}{2019 {Evolutionary}
  {Algorithms} {Review}},'' in \emph{\BIBforeignlanguage{en}{Genetic
  {Programming} {Theory} and {Practice} {XVII}}}, ser. Genetic and
  {Evolutionary} {Computation}, W.~Banzhaf, E.~Goodman, L.~Sheneman,
  L.~Trujillo, and B.~Worzel, Eds.\hskip 1em plus 0.5em minus 0.4em\relax Cham:
  Springer International Publishing, 2020, pp. 307--344.

\bibitem{colledanchise_learning_2019}
M.~Colledanchise, R.~Parasuraman, and P.~{\"O}gren, ``Learning of {{Behavior
  Trees}} for {{Autonomous Agents}},'' \emph{IEEE Transactions on Games},
  vol.~11, no.~2, pp. 183--189, June 2019.

\bibitem{iovino2021learning}
M.~Iovino, J.~Styrud, P.~Falco, and C.~Smith, ``Learning behavior trees with
  genetic programming in unpredictable environments,'' in \emph{2021 IEEE
  International Conference on Robotics and Automation (ICRA)}.\hskip 1em plus
  0.5em minus 0.4em\relax IEEE, 2021, pp. 4591--4597.

\bibitem{iovino2023framework}
------, ``A framework for learning behavior trees in collaborative robotic
  applications,'' \emph{2023 IEEE International Conference on Automation
  Science and Engineering (CASE)}, 2023.

\bibitem{mayr2022combining}
M.~Mayr, F.~Ahmad, K.~Chatzilygeroudis, L.~Nardi, and V.~Krueger, ``Combining
  planning, reasoning and reinforcement learning to solve industrial robot
  tasks,'' \emph{IROS 2022 Workshop on Trends and Advances in Machine Learning
  and Automated Reasoning for Intelligent Robots and Systems}, 2022.

\bibitem{mayr22priors}
M.~Mayr, C.~Hvarfner, K.~Chatzilygeroudis, L.~Nardi, and V.~Krueger, ``Learning
  skill-based industrial robot tasks with user priors,'' in \emph{2022 IEEE
  18th International Conference on Automation Science and Engineering
  (CASE)}.\hskip 1em plus 0.5em minus 0.4em\relax IEEE, 2022, pp. 1485--1492.

\bibitem{ahmad2023learning}
F.~Ahmad, M.~Mayr, and V.~Krueger, ``Learning to adapt the parameters of
  behavior trees and motion generators to task variations,'' in \emph{2023
  IEEE/RSJ International Conference on Intelligent Robots and Systems}.\hskip
  1em plus 0.5em minus 0.4em\relax IEEE, 2023.

\bibitem{colledanchise_towards_2019}
M.~Colledanchise, D.~Almeida, and P.~{\"O}gren, ``Towards {{Blended Reactive
  Planning}} and {{Acting}} using {{Behavior Trees}},'' in \emph{2019
  {{International Conference}} on {{Robotics}} and {{Automation}} ({{ICRA}})},
  May 2019, pp. 8839--8845.

\bibitem{gustavsson_combining_2022}
O.~Gustavsson, M.~Iovino, J.~Styrud, and C.~Smith, ``Combining {{Context
  Awareness}} and {{Planning}} to {{Learn Behavior Trees}} from
  {{Demonstration}},'' in \emph{2022 31st {{IEEE International Conference}} on
  {{Robot}} and {{Human Interactive Communication}} ({{RO-MAN}})}, Aug. 2022,
  pp. 1153--1160.

\bibitem{tumova_maximally_2014}
\BIBentryALTinterwordspacing
J.~Tumova, A.~Marzinotto, D.~V. Dimarogonas, and D.~Kragic,
  ``\BIBforeignlanguage{en}{Maximally satisfying {LTL} action planning},'' in
  \emph{\BIBforeignlanguage{en}{2014 {IEEE}/{RSJ} {International} {Conference}
  on {Intelligent} {Robots} and {Systems}}}.\hskip 1em plus 0.5em minus
  0.4em\relax Chicago, IL, USA: IEEE, Sept. 2014, pp. 1503--1510. [Online].
  Available: \url{http://ieeexplore.ieee.org/document/6942755/}
\BIBentrySTDinterwordspacing

\bibitem{colledanchise_synthesis_2017}
M.~Colledanchise, R.~M. Murray, and P.~Ögren, ``Synthesis of
  correct-by-construction behavior trees,'' in \emph{2017 {IEEE}/{RSJ}
  {International} {Conference} on {Intelligent} {Robots} and {Systems}
  ({IROS})}, Sept. 2017, pp. 6039--6046.

\bibitem{holzl_reasoning_2015}
M.~Hölzl and T.~Gabor, ``\BIBforeignlanguage{en}{Reasoning and {Learning} for
  {Awareness} and {Adaptation}},'' in \emph{\BIBforeignlanguage{en}{Software
  {Engineering} for {Collective} {Autonomic} {Systems}: {The} {ASCENS}
  {Approach}}}, ser. Lecture {Notes} in {Computer} {Science}, M.~Wirsing,
  M.~Hölzl, N.~Koch, and P.~Mayer, Eds.\hskip 1em plus 0.5em minus 0.4em\relax
  Cham: Springer International Publishing, 2015, pp. 249--290.

\bibitem{rovida_extended_2017}
F.~Rovida, B.~Grossmann, and V.~Krüger, ``Extended behavior trees for quick
  definition of flexible robotic tasks,'' in \emph{2017 {IEEE}/{RSJ}
  {International} {Conference} on {Intelligent} {Robots} and {Systems}
  ({IROS})}, Sept. 2017, pp. 6793--6800.

\bibitem{masson2016reinforcement}
W.~Masson, P.~Ranchod, and G.~Konidaris, ``Reinforcement learning with
  parameterized actions,'' in \emph{Proceedings of the AAAI Conference on
  Artificial Intelligence}, vol.~30, no.~1, 2016.

\bibitem{dalal2021accelerating}
M.~Dalal, D.~Pathak, and R.~R. Salakhutdinov, ``Accelerating robotic
  reinforcement learning via parameterized action primitives,'' \emph{Advances
  in Neural Information Processing Systems}, vol.~34, pp. 21\,847--21\,859,
  2021.

\bibitem{deisenroth13r}
\BIBentryALTinterwordspacing
M.~P. Deisenroth, G.~Neumann, and J.~Peters, ``A {{Survey}} on {{Policy
  Search}} for {{Robotics}},'' \emph{Foundations and Trends® in Robotics},
  vol.~2, no. 1–2, pp. 1--142, 2013. [Online]. Available:
  \url{https://www.nowpublishers.com/article/Details/ROB-021}
\BIBentrySTDinterwordspacing

\bibitem{chatzilygeroudis2019survey}
K.~Chatzilygeroudis, V.~Vassiliades, F.~Stulp, S.~Calinon, and J.-B. Mouret,
  ``A survey on policy search algorithms for learning robot controllers in a
  handful of trials,'' \emph{IEEE Transactions on Robotics}, vol.~36, no.~2,
  pp. 328--347, 2019.

\bibitem{chatzilygeroudis172iicirsi}
K.~Chatzilygeroudis, R.~Rama, R.~Kaushik, D.~Goepp, V.~Vassiliades, and J.-B.
  Mouret, ``Black-box data-efficient policy search for robotics,'' in
  \emph{2017 {{IEEE}}/{{RSJ International Conference}} on {{Intelligent
  Robots}} and {{Systems}} ({{IROS}})}, 2017, pp. 51--58.

\bibitem{hvarfner2022pi}
C.~Hvarfner, D.~Stoll, A.~Souza, M.~Lindauer, F.~Hutter, and L.~Nardi, ``Pibo:
  Augmenting acquisition functions with user beliefs for bayesian
  optimization,'' \emph{arXiv preprint arXiv:2204.11051}, 2022.

\bibitem{hutter2011sequential}
F.~Hutter, H.~H. Hoos, and K.~Leyton-Brown, ``Sequential model-based
  optimization for general algorithm configuration,'' in \emph{Learning and
  Intelligent Optimization: 5th International Conference, LION 5, Rome, Italy,
  January 17-21, 2011. Selected Papers 5}.\hskip 1em plus 0.5em minus
  0.4em\relax Springer, 2011, pp. 507--523.

\bibitem{geurts2006extremely}
P.~Geurts, D.~Ernst, and L.~Wehenkel, ``Extremely randomized trees,''
  \emph{Machine learning}, vol.~63, pp. 3--42, 2006.

\bibitem{wu2022autotuning}
X.~Wu, \emph{et~al.}, ``Autotuning polybench benchmarks with llvm clang/polly
  loop optimization pragmas using bayesian optimization,'' \emph{Concurrency
  and Computation: Practice and Experience}, vol.~34, no.~20, p. e6683, 2022.

\bibitem{mayr2021learning}
M.~Mayr, K.~Chatzilygeroudis, F.~Ahmad, L.~Nardi, and V.~Krueger, ``Learning of
  parameters in behavior trees for movement skills,'' in \emph{2021 IEEE/RSJ
  International Conference on Intelligent Robots and Systems (IROS)}.\hskip 1em
  plus 0.5em minus 0.4em\relax IEEE, 2021, pp. 7572--7579.

\bibitem{zhu2020robosuite}
Y.~Zhu, J.~Wong, A.~Mandlekar, and R.~Mart{\'\i}n-Mart{\'\i}n, ``robosuite: A
  modular simulation framework and benchmark for robot learning,'' \emph{arXiv
  preprint arXiv:2009.12293}, 2020.

\bibitem{mayr23iros}
M.~Mayr, F.~Rovida, and V.~Krueger, ``Skiros2 - a skill-based robot control
  platform for ros,'' in \emph{2023 {{IEEE}}/{{RSJ International Conference}}
  on {{Intelligent Robots}} and {{Systems}}}, 2023.

\bibitem{mayr2023using}
M.~Mayr, F.~Ahmad, A.~Duerr, and V.~Krueger, ``Using knowledge representation
  and task planning for robot-agnostic skills on the example of contact-rich
  wiping tasks,'' in \emph{2023 IEEE 18th International Conference on
  Automation Science and Engineering (CASE)}.\hskip 1em plus 0.5em minus
  0.4em\relax IEEE, 2023.

\bibitem{sprague2022adding}
C.~I. Sprague and P.~{\"O}gren, ``Adding neural network controllers to behavior
  trees without destroying performance guarantees,'' in \emph{2022 IEEE 61st
  Conference on Decision and Control (CDC)}.\hskip 1em plus 0.5em minus
  0.4em\relax IEEE, 2022, pp. 3989--3996.

\end{thebibliography}

\end{document}